\documentclass[runningheads]{llncs}
\usepackage{graphicx}
\usepackage{cleveref}
\usepackage{color} 

\begin{document}
\title{Optimizing the Procedure of CT Segmentation Labeling}
%
%
\author{Yaroslav Zharov\inst{1,3} \and
Tilo Baumbach\inst{1,2} \and
Vincent Heuveline\inst{3}}
\authorrunning{Y. Zharov et al.}
%
\institute{Laboratory for Applications of Synchrotron Radiation (LAS), KIT, Germany \and
Institute for Photon Science and Synchrotron Radiation (IPS), KIT, Germany \and
Engineering Mathematics and Computing Lab (EMCL), Heidelberg University, Germany}

%
\maketitle              
\begin{abstract}
In Computed Tomography, machine learning is often used for automated data processing. However, increasing model complexity is accompanied by increasingly large volume datasets, which in turn increases the cost of model training.
Unlike most work that mitigates this by advancing model architectures and training algorithms, we consider the annotation procedure and its effect on the model performance.
We assume three main virtues of a good dataset collected for a model training to be \emph{label quality}, \emph{diversity}, and \emph{completeness}.
We compare the effects of those virtues on the model performance using open medical CT datasets and conclude, that \emph{quality} is more important than \emph{diversity} early during labeling; the \emph{diversity}, in turn, is more important than \emph{completeness}.
Based on this conclusion and additional experiments, we propose a labeling procedure for the segmentation of tomographic images to minimize efforts spent on labeling while maximizing the model performance.
\keywords{Segmentation  \and Computed Tomography \and Labeling.}
\end{abstract}
\section{Introduction}

Neural Networks for computer vision has achieved results on par with human experts \cite{Avetisian2020}.
However, these models require large datasets to be trained.
Which leads to constant pressure to collect and label more training data.
The bottleneck of collecting and sharing data becomes less prominent with increased automation, cheaper storage, and faster networks.
Although, labeling becomes more expensive as the data resolution grows and tasks become more complex.
Subject-matter experts (e.g., medical practitioners) are, therefore, required to spend many hours as labeling experts at the cost of other duties.

Multiple vectors of research are focused on the question, of how to minimize the amount of labeled data required for training: transfer learning \cite{Zhuang2019}, self-supervised learning \cite{Jing2019}, active learning \cite{Ren2022}, etc.
While only active learning modifies the labeling procedure, others mostly assume the dataset given.
In application areas of Computed Tomography (CT), the number of data sets is rapidly increasing due to advances in instrumentation and efficient digital pixel array detectors \cite{Kiryati2021}.
Due to the diversity of biological or technical specimens, and due to strict data requirements of the medical area (e.g., anonymity and disease representation), task-specific datasets are often collected.
To help the community meet this growing demand for datasets, we consider the question of how to acquire the best dataset with minimal effort.
To define what the best dataset is, we note that the dataset collection is not the end of the process, but rather an intermediate step to produce a good model.
The value of the dataset collected to train a model, therefore, can be measured as the performance of a model trained on this dataset.
We focus this paper on medical CT segmentation, however, our conclusions hold for other CT segmentation tasks, since we never impose assumptions that are specific to the medical imaging data.

We assume, that the main virtues of a good dataset collected for training are \emph{quality}, \emph{diversity}, and \emph{completeness}.
The \emph{quality} itself could be separated into the data quality and the label quality.
In case of the data collected for model training, the data quality is defined by the application of the model.
We will, therefore, exclude it from the consideration.
The labeling quality is defined by the imperfectness of the annotation \cite{Losel2020}, where the segmentation labels are not accurately aligned with the actual anatomy presented in the images.
Hereinafter the \emph{quality} refers to the label quality.
The \emph{diversity} is understood as the ability to represent the general variety of samples controlled by known parameters (e.g., patient's age, sex, and anamnesis).
We understand the \emph{completeness} as the ability to represent the natural fluctuations of the human morphology (e.g., even twins can have slightly different morphological features).
From the point of view of the data manifold, \emph{diversity} is the ability to sparsely cover the whole manifold with representative prototypical examples. 
In contrast, \emph{completeness} is the ability to populate it densely and represent the distribution nuances.

In this work, we want to identify the relative importance of these virtues for model performance.
If we have all three virtues fulfilled to the maximal extent in a dataset, no model is required, since any possible example is already incorporated to a dataset, and prediction function is just a lookup table.
In practice, however, we train models to interpolate the full manifold from sparse points provided in the training dataset.
Of course, improving a dataset by each virtue makes a better model.
However, given the limited time, experts balance them guided by intuition and often even implicitly.

In contrast to the concurrent work \cite{Kim2022}, we focus on the segmentation task specifically, and consider only the fully supervised training, as opposed to the weak supervision as a way to vary label granularity.
We also propose a labeling procedure that optimizes the effort.

\section{Method}
\label{sec:setup}

\subsection{Datasets Preparation \& Model Training}
In this study we choose the brain tumor, heart, and liver tasks from the Medical Decathlon segmentation datasets \cite{Antonelli2021}.
For the sake of simplicity, we joined all available classes, to represent binary segmentation, however, our private experience shows that the results hold for a multiclass segmentation.
For each dataset, we took the openly available markup and split substracted 20\% of the available data as the \texttt{test} set.
The \texttt{test} is selected once for a dataset and never altered.
We other 80\% \texttt{train+val} set, as it will be split again later on.

In medical datasets, it is typical to have a relatively small amount of volumes collected from a representative variety of patients \cite{Luca2022}.
Hence, we assume that the portion of random volumes used for training could be a proxy to the \emph{diversity}, and we specify it as a number $\in (0,1]$ representing this portion.
Although this subsampling also affects \emph{completeness}, as we show in \Cref{sec:dvc} and especially in \Cref{fig:d-c-importance}, the model responds differently to diversity and completeness.
We conclude that this is a plausible and sufficient proxy for the purposes of qualitative comparisons presented in this paper.
We follow \cite{Zettler2021}, and always train a 2D model on slices, instead of a 3D one on volumes.
Based on assumption that adjacent slices represent small variations of roughly the same morphology, we use the portion of the slices used for training as a proxy for the \emph{completeness}, which is reported as number $\in (0,1]$ as well.
Finally, as a proxy for the \emph{quality} of the dataset, we take a subset of equidistant slices and interpolate the labels between them using the nearest neighbor approach.
Varying the distance between slices, and therefore the interpolation errors, allows us to manipulate the label \emph{quality}, which we report as a percent $\in [0,100]$ representing the IoU between the label after interpolation and the original label.

To measure the model performance for some virtue value, we:
\begin{enumerate}
    \item modify the \texttt{train+val} part of the dataset to model some virtue:
        \begin{itemize}
            \item sample a portion of volumes to model \emph{diversity};
            \item sample a portion of slices containing a mask to model \emph{completeness};
            \item sample an equidistant set of slices and interpolate markup between them to model \emph{quality}.
        \end{itemize}
    \item split the resulting data into \texttt{train} and \texttt{val}, at a ratio of 80 to 20.
    \item upsample the \texttt{train} in such a way, that the amount of labeled slices is always equal to 80\% of labeled slices in the original \texttt{train+val} set;
    \item fit the model on \texttt{train}, select the best snapshot on \texttt{val}, and measure the model quality on \texttt{test}.
\end{enumerate}

We hypothesized that, while tuning the model and optimizer hyperparameters can change the model performance, it will not change the relative importance of different dataset virtues for the model performance.
Therefore, we always train the same model (UNet \cite{Ronneberger2015} with ResNet-18 \cite{He2016a} as the backbone), with the same optimizer (Adam \cite{Kingma2015} with $3e-4$ learning rate), for the same amount of epochs (100 epochs and 10 epochs long cooldown of the early stopping).
For each measurement, the median of 5 runs on random \texttt{train+val} splits is reported.

\subsection{Results Interpretation}
The target of optimization of the labeling procedure is to obtain the model with the best performance given a certain available effort budget for labeling.
For example, an expert can roughly segment 10 volumes, or spend the same time, precisely segmenting 3 volumes.
For experiments, we devise custom proxies of the effort measure.
We leave the empirical measurement of the effort (e.g., as elapsed time) for future research, however, we consulted with experts involved in the segmentation process, and they concur with our estimation.

To find the optimal strategy we consider a plot, where the horizontal axis describes labeling efforts spent, and the vertical axis represents the model performance (to make the plots clearer we normalize the model performance to the quality of the model trained on the unaltered data).
For the same amount of effort spent pursuing different virtues, we will have different model qualities.
The optimal strategy of labeling is represented by a convex polyline that passes through the points on the plot in such a way that no points lie above it.
Following this trajectory provides the best possible dataset at any given moment.

To understand the optimal strategy, we consider another plot.
On the horizontal axis, we plot the model performance, and on the vertical axes--the value of the compared virtues.
For each point on the optimal trajectory, we add one point per virtue in comparison.
Therefore, each vertically aligned set of points represents the virtues required to achieve a specific model performance.
This way, moving along the horizontal axis, we can see which virtue should be pursued earlier on, to stay on the optimal trajectory.

\section{Results}

\subsection{What is More Important, \emph{Quality} or \emph{Diversity}?}

To compare \emph{quality} and \emph{diversity}, we define effort as the portion of the volumes used (as a measure of \emph{diversity}) multiplied by \emph{quality}.
E.g., $0.1$ of the volumes segmented with $80\%$ IoU will result in $8\%$ effort.

\begin{figure}
  \centering
  \includegraphics[width=\textwidth,keepaspectratio]{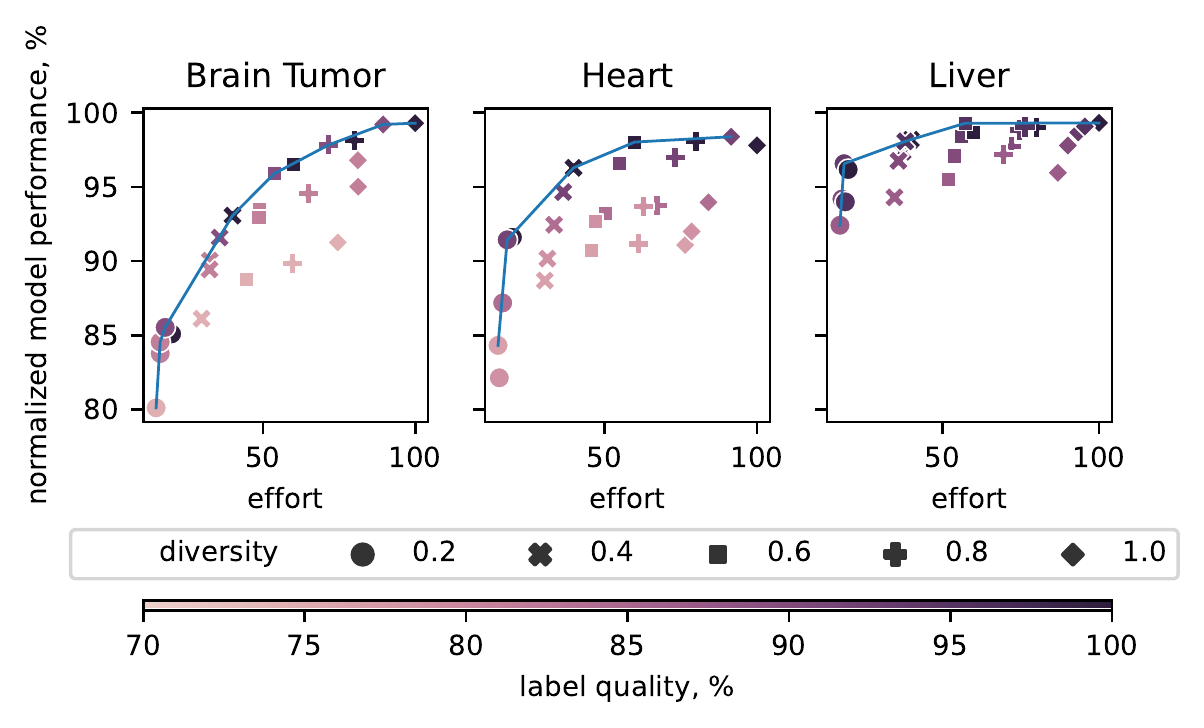}
  \caption{Optimal trajectory w.r.t. \emph{quality} and \emph{diversity}. The variation of the \emph{quality} is represented by color, of \emph{diversity} by the marker shape.}
  \label{fig:q-d-hull}
\end{figure}

The sampled plot with the optimal trajectory is shown in \Cref{fig:q-d-hull}.
From this plot, we observe that the optimal trajectory connects the high-quality points, while low-quality points always fall far below the line.

\begin{figure}
  \centering
  \includegraphics[width=\textwidth,height=0.75\textheight,keepaspectratio]{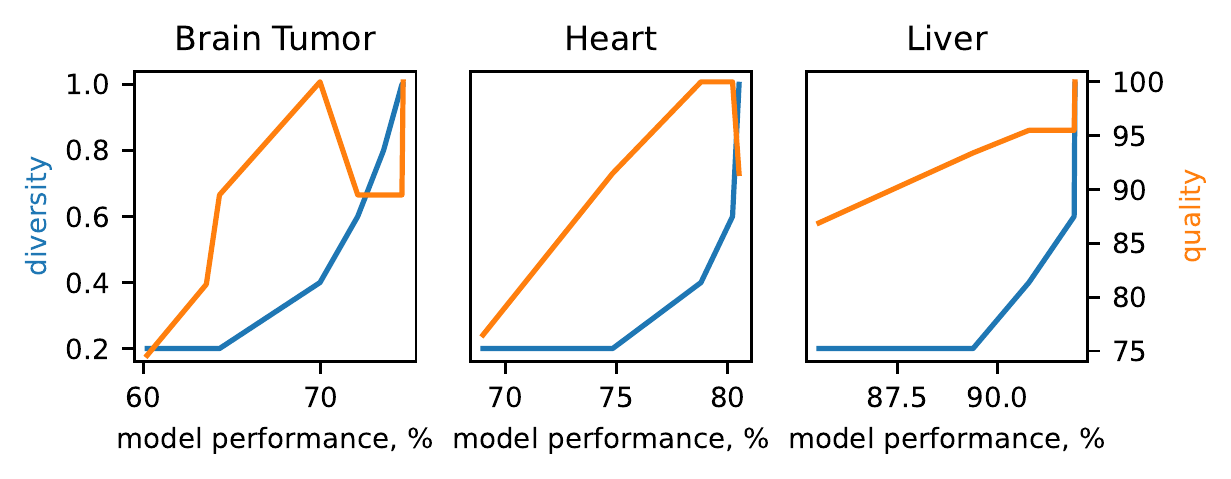}
  \caption{Importance of the \emph{diversity} and \emph{quality}. The orange line represents label \emph{quality}, and blue--\emph{diversity}.}
  \label{fig:q-d-importance}
\end{figure}

We show in \Cref{fig:q-d-importance} how \emph{quality} and \emph{diversity} drive the optimal trajectory.
From this plot we conclude, that \emph{quality} is more important early on, even though we never use IoU worse than 75\% (which could be admissible quality for small area labels).
However, as \emph{quality} reaches around 90\%, increasing \emph{diversity} becomes as important or even more important than increasing \emph{quality}.

\subsection{How Much Labeling \emph{Quality} is Enough?}

\begin{figure}
  \centering
  \includegraphics[width=\textwidth,height=0.75\textheight,keepaspectratio]{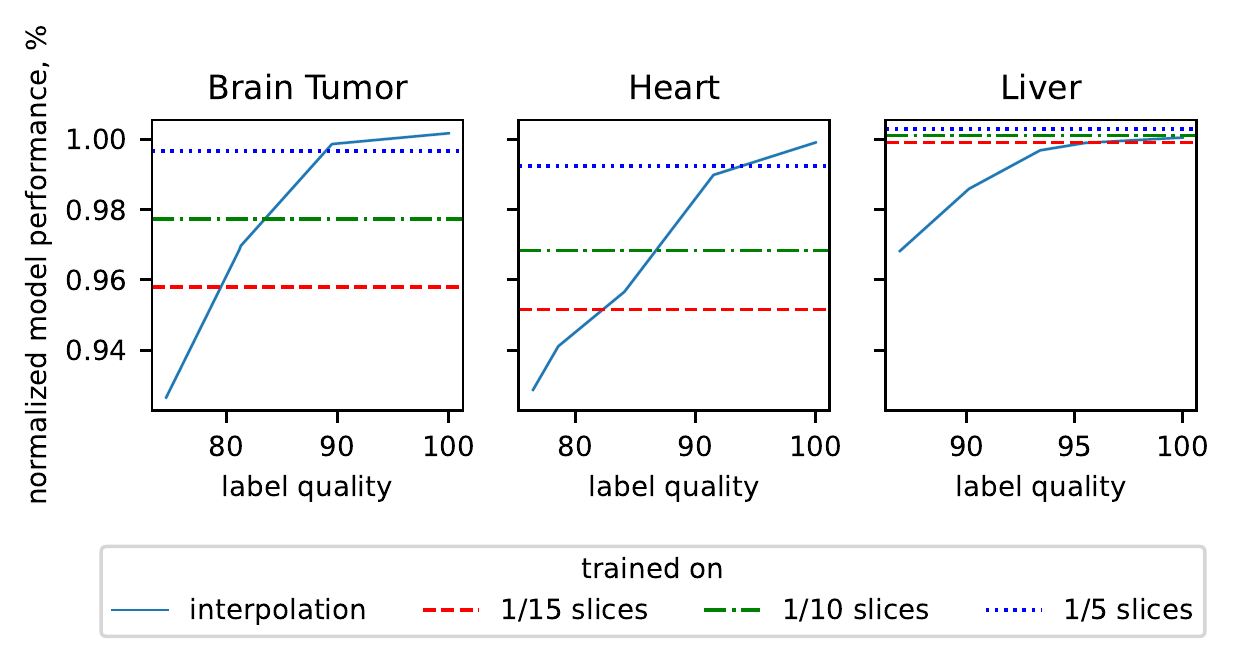}
  \caption{Saturation of the model performance with the increase of the label \emph{quality}. Additionally, the performance of the model trained on a small portion of $1.0$ quality slices is represented with horizontal lines.}
  \label{fig:q-q}
\end{figure}

Increasing labeling quality up to 100\% is challenging if not impossible.
But where is a meaningful threshold of the labeling quality, after which the model performance stagnates?
To investigate it, we plot the model performance against the labeling quality (see \Cref{fig:q-q}).
We also plot the performance of the model trained on sparsely segmented slices (each 5th, 10th, and 15th).
We conclude, that, first, if one can not achieve good interpolation quality, it may be even harmful to interpolate, and, second, in accordance with the previous section--one should aim for 90\% quality of the labeling before aiming for either \emph{completeness} or \emph{diversity}.

\subsection{What is More Important, \emph{Diversity} or \emph{Completeness}?}
\label{sec:dvc}

To compare \emph{diversity} and \emph{completeness}, we define the effort as a total percentage of slices segmented. 
E.g., if we sample 6 volumes from 10 available (\emph{diversity} = $0.6$) and segmented each 10th slice (\emph{completeness} = $0.1$), then the effort is $0.06$.

\begin{figure}
  \centering
  \includegraphics[width=\textwidth,height=0.75\textheight,keepaspectratio]{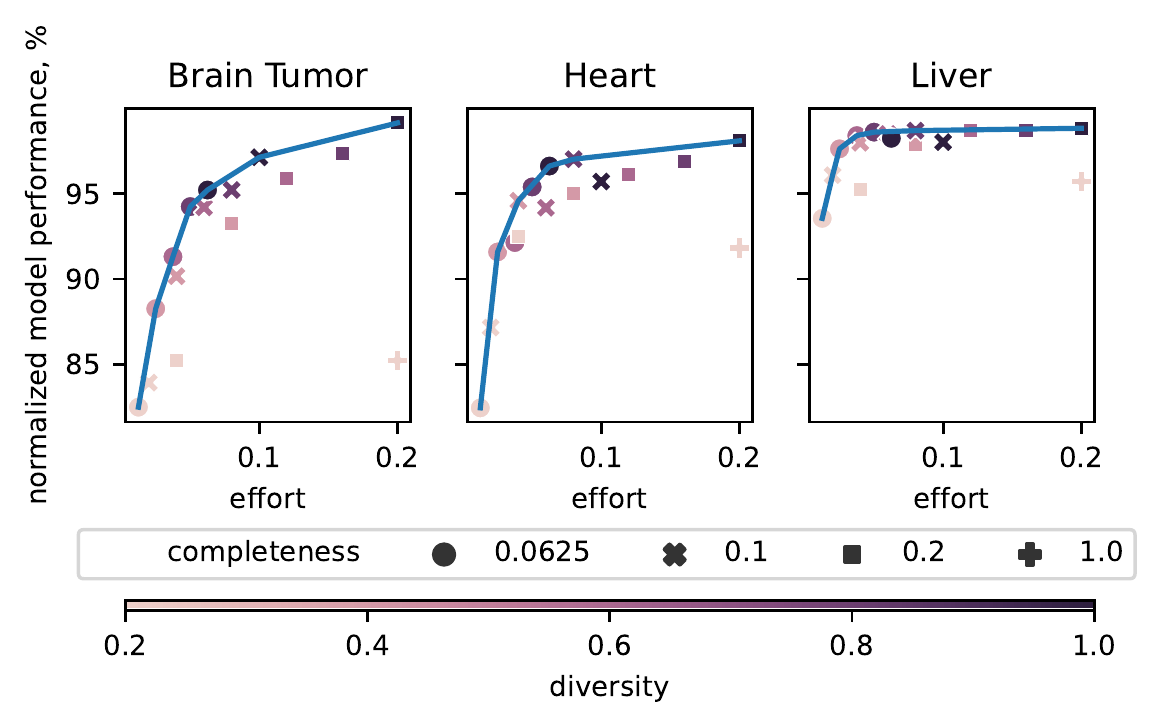}
  \caption{Optimal trajectory w.r.t. \emph{completeness} and \emph{diversity}. Variation of \emph{diversity} is represented by color, of \emph{completeness} by the marker shape.}
  \label{fig:d-c-hull}
\end{figure}

The optimal trajectory plot is shown in \Cref{fig:d-c-hull}.
The brightest demonstration of the importance of \emph{diversity} is in the right bottom parts: the $0.2$ of \emph{diversity} with $1$ of \emph{completeness} is much worse than vice versa.

\begin{figure}
  \centering
  \includegraphics[width=\textwidth,height=0.75\textheight,keepaspectratio]{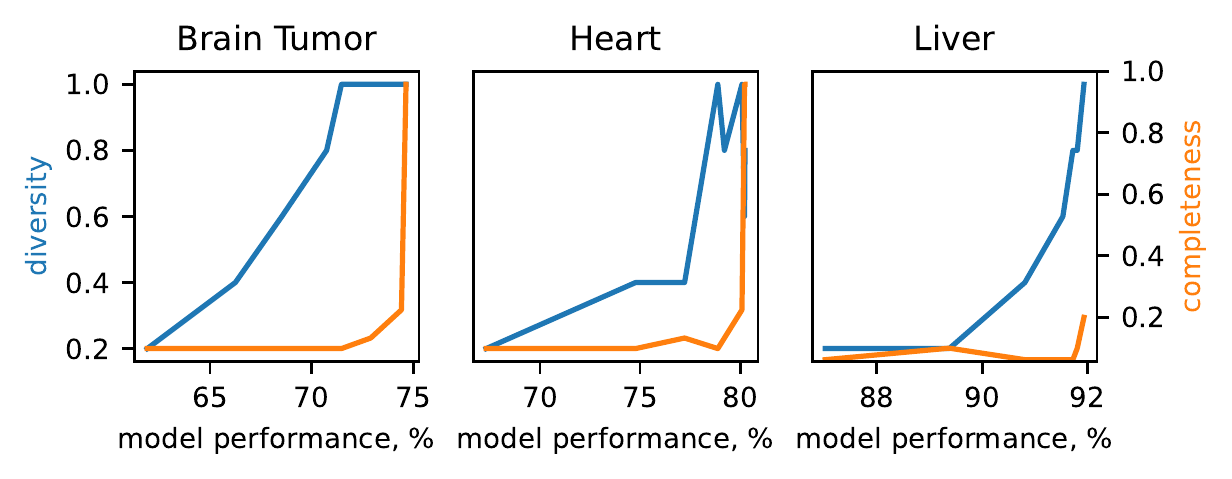}
  \caption{Importance of the \emph{diversity} and \emph{completeness}. The orange line represents label \emph{completeness}, and blue--\emph{diversity}.}
  \label{fig:d-c-importance}
\end{figure}

In \Cref{fig:d-c-importance} we demonstrate the importance of \emph{diversity} and \emph{completeness} for the optimal trajectory.
Not only \emph{diversity} is more important early on, but \emph{completeness} contributes less to the model performance (note the steeper growth of the \emph{completeness} with the performance growth).
Therefore, finding more diverse examples and segmenting more volumes should be preferred to segmenting more random variations and segmenting slices more densely/interpolating them.

\subsection{How Much Data \emph{Diversity} is Enough?}
\begin{figure}
  \centering
  \includegraphics[width=\textwidth,height=0.75\textheight,keepaspectratio]{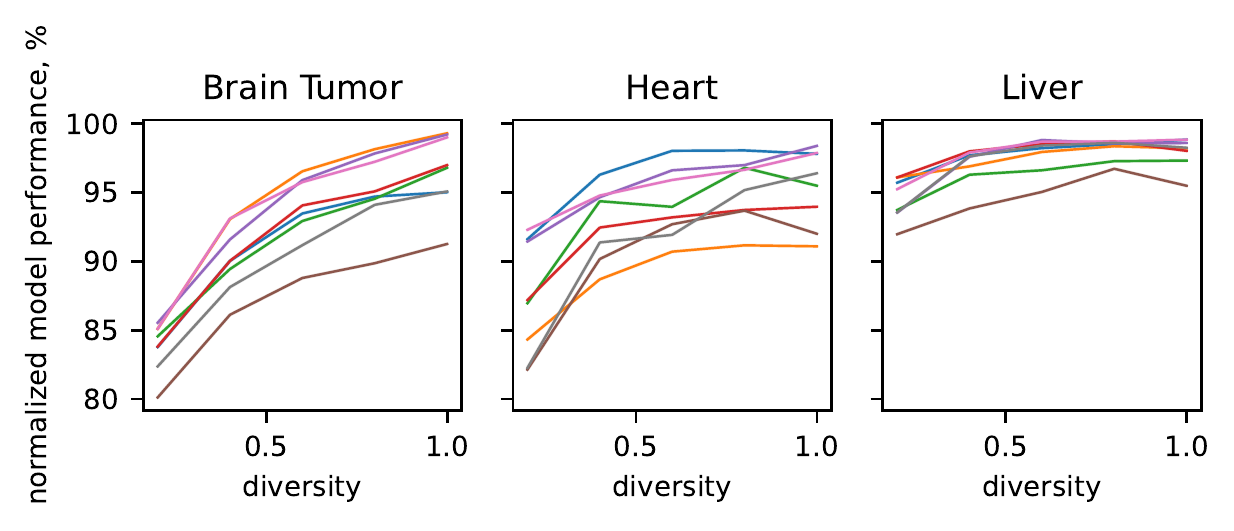}
  \caption{Saturation of the model performance with the increase of the \emph{diversity}. No matter how other virtues vary (different colors), saturation comes at the same point.}
  \label{fig:saturation}
\end{figure}

Although, how to define a reasonable limit to stop searching for \emph{diversity}, and start increasing \emph{completeness}?
We plot the performance of the model w.r.t. \emph{diversity} in \Cref{fig:saturation}, each line represents a different set of virtues.
Since the total amount of the data possibly available is unknown, we can not define numerical limit.
Instead, we note that all lines saturate ca. at the same point of increasing diversity.
Hence we can define where to stop increasing \emph{diversity}, and start \emph{increasing} completeness, by continuously updating a segmentation model, while expanding the dataset.

\section{Conclusion}

In this paper, we have compared the importance of different ways to spend labeling efforts and presented a way to optimize the segmentation labeling procedure.
We minimized the effort required to obtain the model of a specific quality.
In general, we conclude that \emph{quality} is more important than \emph{diversity}, which is more important than \emph{completeness}.
Based on our experiments, we propose the following procedure to minimize the effort during labeling volumetric data for segmentation:

\begin{enumerate}
    \item Start with segmenting slices, without interpolation.
    Aim for maximal quality affordable without pixel hunting, at least 90\%.
    \item Decide on your time budget and distribute slices to segment as evenly through diverse volumes as possible.
    Though, keep in mind, that the structure of interest may impose a minimal slice number per volume to capture all parts of the structure.
    \item Train a model as early in the process as possible. 
    This allows, first, deciding which areas require more markup (by means of active learning, or just by an expert assessment of predictions), and, second, recognizing the moment when model performance starts to saturate w.r.t. \emph{diversity}.
    \item After hitting the saturation w.r.t. the \emph{diversity}, increase the \emph{completeness} either by adding more volumes or by interpolating more slices to squeeze the last performance percent.
\end{enumerate}

We leave a rigorous study of the exact conversion between theoretical effort metrics presented in this paper and empirical efforts (e.g., time spent) for future research.



\subsubsection*{Acknowledgment.} Data used in this paper was published in \cite{Antonelli2021} and is available under a Creative Commons license CC-BY-SA4.0.

\newpage
\bibliographystyle{splncs04}
\bibliography{references}

\end{document}